%% file: main.tex
\documentclass[10pt,twocolumn,letterpaper]{article}

\usepackage{multirow}
\usepackage[pagenumbers]{cvpr} 

\input{preamble}

\definecolor{cvprblue}{rgb}{0.21,0.49,0.74}
\usepackage[pagebackref,breaklinks,colorlinks,citecolor=cvprblue]{hyperref}
\usepackage{multirow} 

\def\paperID{} 
\def\confName{CVPR}
\def\confYear{2024}

\title{Progressive Evolution from Single-Point to Polygon for Scene Text}

\author{
Linger Deng\textsuperscript{1}$^\dag$
\quad Mingxin Huang\textsuperscript{2}$^\dag$
\quad Xudong Xie\textsuperscript{1}
\quad Yuliang Liu\textsuperscript{1}$^*$
\quad Lianwen Jin\textsuperscript{2}
\quad Xiang Bai\textsuperscript{1}
\\
\textsuperscript{1}{Huazhong University of Science and Technology}
\quad \textsuperscript{2}{South China University of Technology} 
\\
{\tt\small lingerdeng2023@163.com \quad ylliu@hust.edu.cn
      }
}

\begin{document}
\maketitle

\begin{abstract}
The advancement of text shape representations towards compactness has enhanced text detection and spotting performance, but at a high annotation cost. Current models use single-point annotations to reduce costs, yet they lack sufficient localization information for downstream applications.
To overcome this limitation, we introduce Point2Polygon, which can efficiently transform single-points into compact polygons.
Our method uses a coarse-to-fine process, starting with creating and selecting anchor points based on recognition confidence, then vertically and horizontally refining the polygon using recognition information to optimize its shape.
We demonstrate the accuracy of the generated polygons through extensive experiments: 1) By creating polygons from ground truth points, we achieved an accuracy of 82.0\% on ICDAR 2015; 2) In training detectors with polygons generated by our method, we attained 86\% of the accuracy relative to training with ground truth (GT); 3) Additionally, the proposed Point2Polygon can be seamlessly integrated to empower single-point spotters to generate polygons. This integration led to an impressive 82.5\% accuracy for the generated polygons. It is worth mentioning that our method relies solely on synthetic recognition information, eliminating the need for any manual annotation beyond single points. 
\end{abstract}
\let\thefootnote\relax\footnotetext{$^\dag$Equal contribution.}
\let\thefootnote\relax\footnotetext{$^*$Corresponding author.}

\section{Introduction}
\label{sec:intro}

In recent years, the evolution of text shape representation has significantly enhanced text detection and spotting, progressing from managing horizontal and multi-oriented text to adeptly handling arbitrarily shaped text. However, polygonal annotation brings expensive annotation costs and limits the acquisition of large-scale annotated data, constraining the overall generalization of the model.

To reduce the requirement of costly annotation, some researchers have delved into weakly supervised methods. Several approaches  \cite{hu2017wordsup,tian2017wetext,wupolygon,baek2019character,zhao2023texts} employ coarse detection annotations to yield refined detection outcomes. Another strategy \cite{kittenplon2022towards} combines synthetic data with box annotations and real data with recognition annotations. Recent research by Liu \textit{et al.} \cite{liu2023spts} demonstrated that employing cost-effective single-point annotations for location supervision can yield competitive results in text spotting. Nonetheless, this approach has limitations in certain downstream applications. Kil \textit{et al.} \cite{kil2023towards} argue that single-point annotations fall short in providing adequate text location information for tasks like scene text editing ~\cite{wu2019editing,qu2023exploring} and text removal ~\cite{liu2020erasenet,wang2021pert}. 

\begin{figure}[t!]
    \centering
    \includegraphics[width=0.9\linewidth]{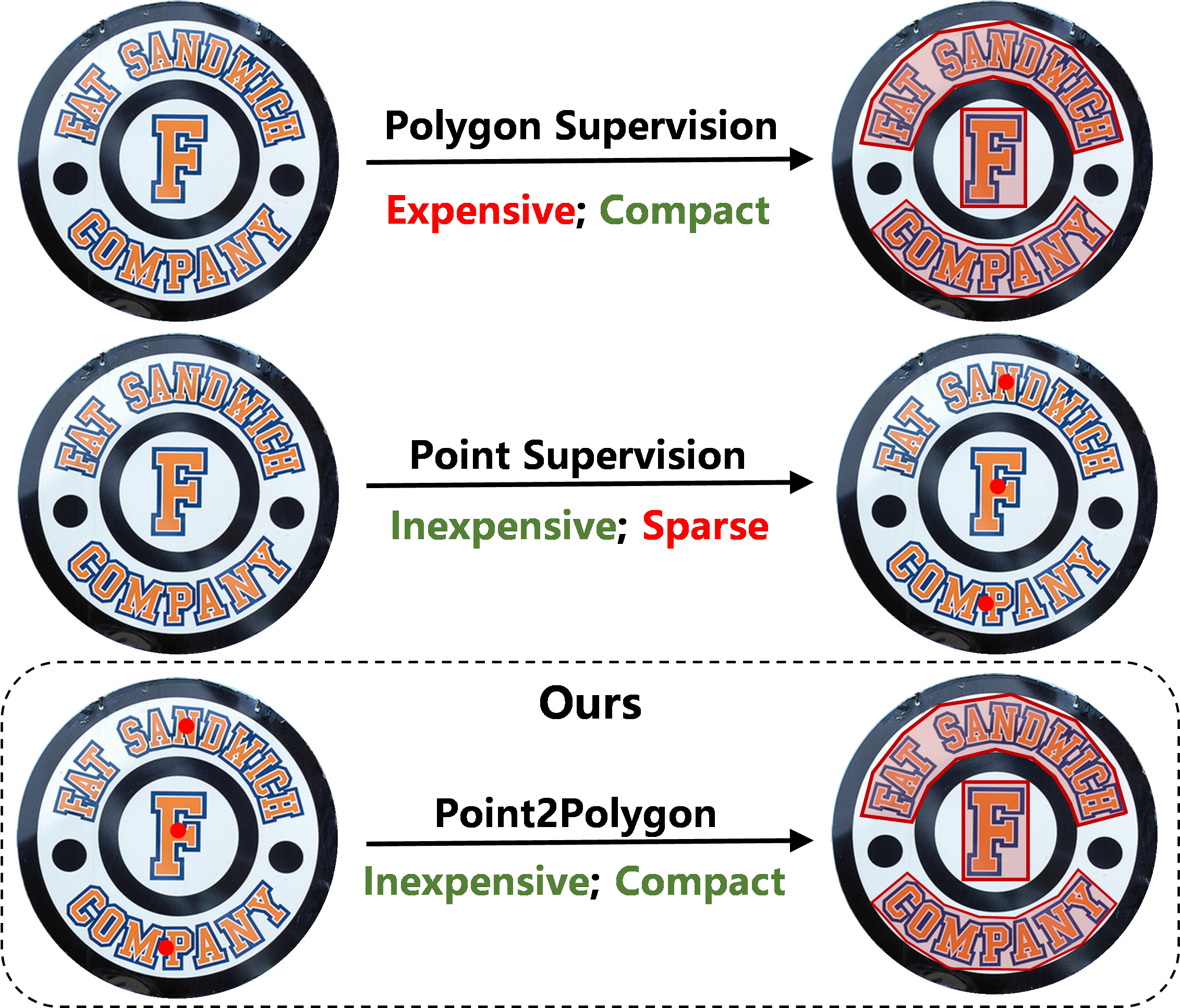}
    \caption{Point2Polygon maintains cost-efficient point annotation while automatically generating polygons with high accuracy.}
    \label{fig:intro}
\end{figure}

    Hence, given the high cost of polygonal annotation and the necessity for accurate text representation, a key question arises, as shown in Fig.~\ref{fig:intro}: \textit{Can we generate text polygon representations solely through single-point annotation, to keep both the advantages of cost-effective annotation and compact representation?}
    
    
    However, deriving an accurate polygon directly from weak supervision of single-point could be a significant challenge due to the considerable variations in shape, size, style, distortion, and layout inherent in scene text. However, recent studies~\cite{shi2018aster,cheng2018aon,huang2022swintextspotter,huang2023estextspotter,ye2023deepsolo} suggest recognition information can be beneficial for predicting accurate detection results. 

    In this paper, we present Point2Polygon, an efficient method to transform single points into polygons using a three-step, coarse-to-fine process. Initially, our Anchor Generation Module (AGM) creates multiple anchors from input image sizes and selects the best one using initial recognition confidence. Next, the Polygon Generation Module (PGM) shapes these anchors into vertical boundary points, guided by recognition loss and supervised by the Thin-Plate Spline technique. The final step involves our Polygon Rectification Module (PRM), which adjusts the polygon horizontally, using detailed recognition information to remove excess points beyond the region of the text. 
    Extensive experiments demonstrate the accuracy of the generated polygons. We summarize the advantages of the proposed method as follows:

    \begin{itemize}
        \item We introduce a novel framework, termed Point2Polygon, which can efficiently
         leverage multi-granularity recognition information to evolve points into polygons. 
        \item Point2Polygon requires only single-point annotations and synthetic recognition data. Its simplicity also allows it to be seamlessly integrated with single-point spotters like SPTS v2 for autonomous polygon generation, with an impressive 82.5\% accuracy.
        \item Experiments demonstrate the accuracy of the generated polygon through multiple aspects: It achieves 82.0\% accuracy on ICDAR 2015 by generating polygons from ground truth points, and an average of 86\% accuracy training using several detectors.
    \end{itemize}

\section{Related Work}

\begin{figure*}[t!]
    \centering
    \includegraphics[width=0.95\linewidth]{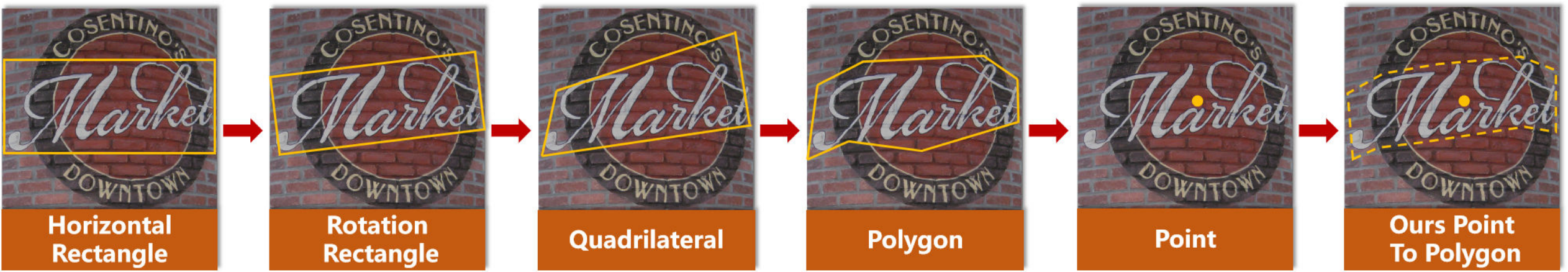}
    \caption{Adapting to the development in text shape representation, we present Point2Polygon, a novel method that significantly alleviates the constraints of point supervision.}
    \label{fig:related}
\end{figure*}

In recent years, text representation of scene text has progressed through distinct stages of development, as presented in Fig.~\ref{fig:related}. Early methods mainly focused on horizontal text detection. For instance, Tian \textit{et al.}~\cite{tian2016detecting} introduced a CTPN that predicts a sequence of fixed-width text fragments along with a vertical anchor frame.
Liao \textit{et al.}~\cite{liao2017textboxes} accurately adjust the anchor box and classification layer based on the SSD~\cite{liu2016ssd}.

Subsequently, the research focus has shifted towards multi-oriented text detection. 
For example, Zhou \textit{et al.}~\cite{zhou2017east} proposed EAST to regress the four points in multi-oriented text.
He \textit{et al.}~\cite{he2017deep} proposed a framework to directly regress the multi-oriented text.

Although significant progress has been achieved, researchers find the previous methods still fall short in handling arbitrarily shaped text. 
To address this issue, two categories have been developed. The first treats text detection as a segmentation task, initially segmenting the text kernel and then expanding to the full text. For example, Wang \textit{et al.}~\cite{wang2019shape} proposed PSENet, which progressively expands the kernel to the final text boundary. For real-time detection, Wang \textit{et al.}~\cite{wang2019efficient} proposed a learnable post-processing method to expand the kernel. Liao \textit{et al.}~\cite{liao2019mask} introduced an adaptive binarization process for the segmentation model. In contrast to segmentation-based methods, regression-based methods utilize curves directly to fit text. Liu \textit{et al.}~\cite{liu2020abcnet} proposed a parameterized Bezier curve to fit the arbitrarily shaped text. Similarly, Liu \textit{et al.}~\cite{liu2023pbformer} adopt polynomials to represent the curve text. 

Although arbitrarily shaped text can precisely represent the shape of the text, the cost of annotation is expensive. Therefore, some researchers have begun to explore reducing the cost of annotation. One solution utilizes the coarse detection annotation to generate fine detection results. Hu \textit{et al.}\cite{hu2017wordsup} and Tian \textit{et al.}\cite{tian2017wetext} have proposed a similar solution, employing word or line-level annotation to achieve character-level detection. Additionally, Wu \textit{et al.}\cite{wupolygon} have utilized curved synthetic data to train a segmentation module tailored for text with arbitrary shapes. 
Another solution involves using synthetic data with box annotation and real data with recognition annotation. Kittenplon \textit{et al.}~\cite{kittenplon2022towards} propose a weakly-supervised method that uses synthetic data with box annotations and real data with text annotations to train the text spotting model. Despite these advancements, it is important to note that both approaches still require the expensive annotation of bounding boxes. 

Therefore, some methods~\cite{peng2022spts,liu2023spts,tang2022you} attempt to further reduce the annotation costs by employing a single point, which has shown significant success in end-to-end text spotting. However, it's important to note that while a single point is effective in end-to-end text spotting, the bounding box remains crucial for numerous downstream tasks, as emphasized in~\cite{kil2023towards}. 
In this paper, we introduce Point2Polygon, a novel approach that generates polygons using only single points and synthetic recognition information.

\section{Methodology}
\begin{figure*}[t!]
    \centering
    \includegraphics[width=0.9\linewidth]{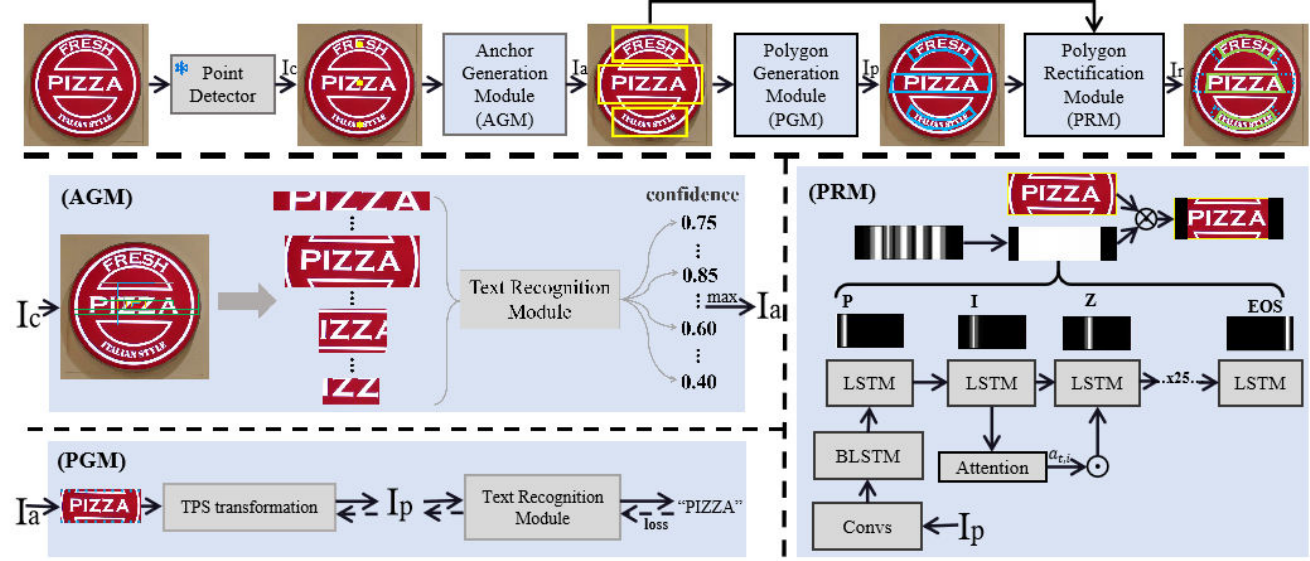}
    \caption{ Overview of the proposed Point2Polygon model. We use the point detector as a basis for obtaining the final text polygon by supervising it from coarse to fine strategy. 
    }
    \label{fig:framework}
\end{figure*}

In this paper, we present a solution for evolving polygons from a single point using coarse-to-fine recognition information. The overall pipeline is illustrated in Fig.~\ref{fig:framework}. Firstly, given an image, a point detector is employed to locate the central point $I_c$ of a text. Then, we introduce an Anchor Generation Module (AGM) to generate adaptive multiple anchors centered on these points and select the optimal one $I_a$. It is selected based on coarse-grained recognition information (recognition confidence). Next, these selected anchors are fed into the Polygon Generation Module (PGM) to refine the upper and lower boundary points $I_p$ using the Thin-Plate spline~\cite{bookstein1989principal}. This module operates under medium-grained recognition information (recognition loss). Finally, since the boundary points obtained in PGM often extend beyond the length of the text horizontally, we introduce the Polygon Rectification Module (PRM). PRM leverages fine-grained recognition information (attention map) to finely adjust the boundary points horizontally, producing a more precise adjustment $I_r$ of the boundary points. The point detector refers to any model capable of accurately pinpointing the central location of textual content solely through point-based supervision.

\subsection{Anchor Generation}
\label{sec:RCGAG}
The Anchor Generation Module (AGM) intercepts multiple anchors centered on the text's central point. Subsequently, these anchors are input into the text recognition module, and the anchor with the highest recognition confidence is selected as the optimal anchor. The Fig.~\ref{fig:framework} illustrates the process of this module.

As the scene text exhibits significant variations in shape, we preset different default anchors to cover them. This includes ``extra-long'' default anchors designed for extra-large aspect ratio text, ``large'' anchors for large aspect ratio text, ``normal'' anchors for moderate aspect ratio text, and ``short'' anchors for small aspect ratio text. Each category is diversified into multiple sizes to effectively accommodate different text instances. We design the preset box according to the average size of the text instance in existing benchmarks. Concretely, we empirically establish 4 sizes for extra-long default anchors, 6 sizes for long default anchors, 5 sizes for normal default anchors, and 6 sizes for short default anchors. Given the image size $\left(w, h \right)$, the sizes of the corresponding anchors are represented as follows:
\begin{align}
 \left(w_{s l}, h_{s l}\right)&=\left(\frac{2 w}{3} \text { or } \frac{2 w}{5}, \frac{h}{5 q}\right),  \\
\left(w_l, h_l\right)&=\left(\frac{2 w}{5 j}; \frac{h}{5 i}\right), \\
\left(w_n, h_n\right)&=\left(\frac{2 w}{5 k}, \frac{2 h}{5 k}\right), \\
\left(w_s, h_s\right)&=\left(\frac{w}{5 i}, \frac{2 h}{5 j}\right),
\end{align}
where $\left(w_{s l}, h_{s l}\right)$, $\left(w_l, h_l\right)$, $\left(w_n, h_n\right)$,  $\left(w_s, h_s\right)$ are extra-long, long, normal, and short default anchors, respectively. $i,j,k,q$ are scaling factors sets, where $q=1,2,3,4$, $i=1,2,...,6$, $j=1,2,4,6,8,10$, and $k=1,2,3,6,10$.

Then, we send the coarse anchors into the off-the-shell text recognition model and obtain the recognition confidence. The recognition confidence of an anchor is the sum of the confidences of each character in the recognition result: $P=\sum_i^n P_i / n$,
where $n$ is the number of characters contained in the text and $P_{i}$ is the prediction confidence for each character.
We select the optimal anchor with the highest recognition confidence.

\subsection{Polygon Generation}
\label{sec:RLPG}
To effectively represent arbitrarily shaped text, we propose a Polygon Generation Module (PGM) to refine the coarse anchors from AGM. The entire procedure is illustrated in the bottom left corner of Fig.~\ref{fig:framework}, labeled as the AGM. Firstly, we evenly sample 10 points along the upper and lower contour for each coarse anchor. Subsequently, we employ the Thin-Plate Spline (TPS) to map these sampled points to the boundaries of the text region. TPS transformation necessitates establishing a correspondence for each coordinate, enabling the transformation from the rectangle to the arbitrary text shape. We refer to the target shape of the text as $\textbf{G}$ and the predefined rectangle as $\textbf{O}$. According to the TPS transform, the corresponding point of $p(x,y) \in \textbf{G}$ on $\textbf{O}$ can be calculated by
\begin{align}
\Phi(p)=m_0+\mathbf{M} \cdot p+\sum_{i=1}^N \omega_i U\left(\left\|p-p_i\right\|\right),
\end{align}
where $M$ and $m_0$ are unknown parameters to be calculated, $N$ is the number of control points, $\left\|p-p_i\right\|$is the distance from $p(x,y)$ to fixed points $p_i(x_i,y_i)$ on the target shape $G$. $U$ is the radial basis function $U(r)= r^2 \ln r$, and $w_i$ denotes the weighting for different radial bases.
Then the TPS transform function is determined by the parameters
\begin{align}
\mathbf{T}=\left[\begin{array}{ccccc}m_0 & M & w_{1} & \ldots & w_{N}\end{array}\right],
\end{align}
whose shape is $[2, N+3]$.

Through the utilization of the TPS transform function, the coarse anchors can be transformed into boundary points that are in closer proximity to the text region, enhancing the precision of localization. Inspired by the text recognition model~\cite{shi2018aster,luo2019moran,ijcai2022p124,zheng2023cdistnet}, we use recognition loss to supervise the learning of the locations of target boundary points on $\textbf{G}$.

\subsection{Polygon Rectification}
We observed that the boundary points generated by the PGM are often excessively wide. To rectify this issue, we introduce a Polygon Rectification Module (PRM). The PRM leverages fine-grained recognition information (attention map), to remove the excess points beyond the region of the text. Following the TPS transformation, this problem does not manifest in the vertical direction. Therefore, we employ a one-dimensional attention map to refine the boundary points from the horizontal direction. The entire procedure is illustrated in the bottom right corner of Fig.~\ref{fig:framework}, labeled as the PRM. We first generate cropped regions by utilizing the boundary points extracted from PGM to crop the image. Subsequently, these cropped regions are forwarded to the PRM, which employs a stack of convolutional layers to extract features from them. Following this, a multi-layer Bidirectional LSTM (BLSTM) network is applied to these features, resulting in a new feature vector with a length of $H=[h_1, ..., h_n]$, where $n$ represents the width of the features. Then, a sequence-to-sequence model is employed to convert the feature sequence into a character sequence, denoted as $(y_1, . , y_T)$. At each step $t$, the sequence-to-sequence model predicts the next character or the end-of-sequence symbol (EOS) based on the features $H$, the internal state $s_{t-1}$, and the symbol $y_{t-1}$ predicted in the previous step. During this process, the sequence-to-sequence model computes a vector of attention weights $a_t$ using an attention mechanism:


\begin{align}
c_{t, i} & =\mathbf{w}^{\top} \tanh \left(\mathbf{W}_{\mathbf{s}_{t-1}}+\mathbf{V h}_i+b\right) \\
a_{t, i} & =\exp \left(c_{t, i}\right) / \sum_{i^{\prime}=1}^n \exp \left(c_{t, i^{\prime}}\right)
\end{align}

Where w, W, V are trainable weights. After obtaining the attention maps of each time step, we use a threshold $\tau$ to transform the attention maps. This process is formalized as:
\begin{align}
\mathbf{a}_{t} = \left\{
\begin{aligned}
1 & , & \mathbf{a}_{t} > \tau ,, \\
0 & , & \mathbf{a}_{t} < \tau ,.
\end{aligned}
\right.
\end{align}
We assign attention weights of 1 to steps where the attention weights surpass a specified threshold $\tau$, and 0 to the rest. The attention map efficiently locates the character positions in the feature map. As a result, with the attention map, we can exclude points with zero attention weight, eliminating unnecessary information.

\section{Experiments}
To evaluate the effectiveness of our method, we conduct experiments on the popular scene text detection benchmarks, including ICDAR 2015~\cite{karatzas2015icdar}, TotalText~\cite{ch2020total}, and SCUT-CTW1500~\cite{liu2019curved}.

\subsection{Implementation Details}
\label{Implementation Details}
In this paper, we employ the SPTS v2 method~\cite{liu2023spts}, which utilizes autoregressive decoding to generate single-point coordinates, as the point detector for assessing the model's performance. We specifically select ASTER~\cite{shi2018aster} as our text recognition model. We utilize the synthetic datasets Synth90k~\cite{jaderberg2014synthetic} and SynthText~\cite{gupta2016synthetic} for training our text recognizer. We employ the Adadelta optimizer~\cite{zeiler2012adadelta} and initially set the learning rate to 1.0, which is later reduced to 0.1 at step 0.6M and further to 0.01 at step 0.8M.

\subsection{Datasets}

\textbf{ICDAR 2015.} This dataset consists of 1000 training images and 500 test images of incidental scenes with complex backgrounds and multidirectional text. Some text can also appear in any orientation and at any location at small or low resolution, with high variability in blurring, distortion, etc. 

\noindent
\textbf{TotalText.} This dataset is designed to be used for arbitrarily-shaped text detection and recognition, which contains 1,255 training images and 300 test images. The images are annotated with word-level annotations.

\noindent
\textbf{SCUT-CTW1500.} This dataset is an arbitrarily-shaped scene text benchmark in line-level annotations. It contains 1000 training images for training and 500 images for testing.

\subsection{Ablation Study}
In this section, we utilize the point detector for text center point prediction to conduct the ablation studies. It is worth noting that the recognizer in this paper is trained in the word-level synthetic recognition data. Therefore, the performance on CTW1500 will be lower than the word-level benchmark ICDAR2015 and TotalText.

\begin{table*}[!t]
\centering
\renewcommand\arraystretch{1.2}
\setlength{\tabcolsep}{2mm}{
\begin{tabular}{cccccccccccc}
\hline
\multicolumn{3}{c}{\multirow{2}{*}{Method Selection}} & \multicolumn{3}{c}{\multirow{2}{*}{ICDAR15}} & \multicolumn{3}{c}{\multirow{2}{*}{TotalText}} & \multicolumn{3}{c}{\multirow{2}{*}{CTW1500}} \\
\multicolumn{3}{c}{}                                  & \multicolumn{3}{c}{}                         & \multicolumn{3}{c}{}                            & \multicolumn{3}{c}{}                         \\ \hline
AGM              & PGM              & PRM             & P           & R    & H      & P    & R   & H    & P            & R     & H     \\ \hline
\checkmark        & \checkmark        & \checkmark       & 77.1     & 70.7  & 73.8   & 72.0      & 71.0   & 71.5  & 54.2     & 56.7  & 55.4        \\
                 & \checkmark        & \checkmark       & 18.5         & 17.1         & 17.8         & 11.8          & 11.5          & 11.6         & 6.6          & 10.3         & 8.1         \\
\checkmark        &                  & \checkmark       & 70.7         & 65.2         & 67.9        & 60.7          & 60.2          & 60.4         & 49.1         & 51.4         & 50.2        \\
\checkmark        & \checkmark        &                 & 74.4         & 68.3         & 71.2        & 71.7          & 70.6          & 71.1         & 54.0         & 56.4          & 55.2        \\ \hline
\end{tabular}
}
\caption{Ablation studies on the proposed designs. AGM means the Anchor Generation Module. PGM means the Polygon Generation Module. PRM means the Polygon Rectification Module. P represents the Precision. R represents the Recall. and H represents the Hmean.}
\label{three_schedule}
\end{table*}

\begin{figure}[!t]
    \centering
    \includegraphics[width=8.cm, height=5.0cm]{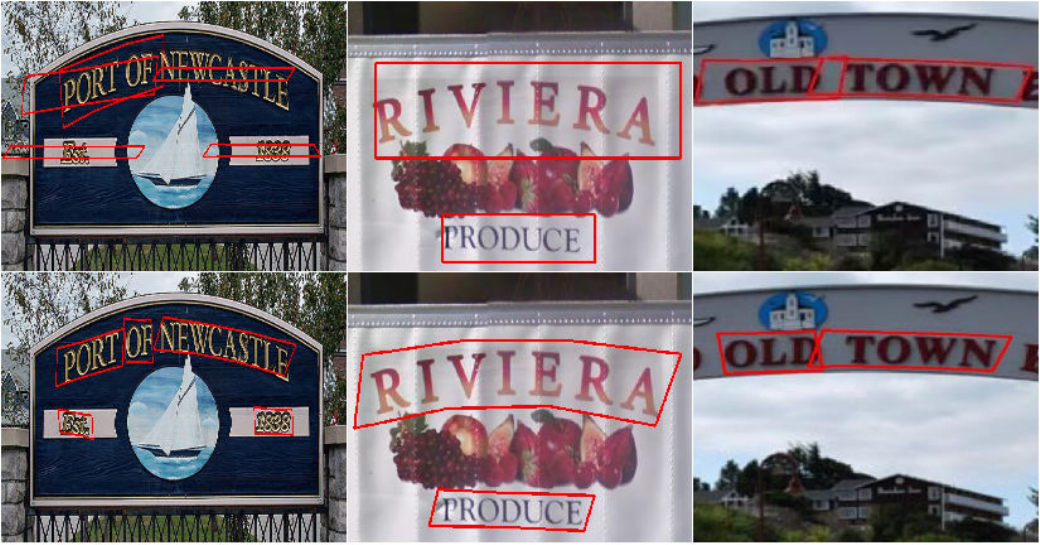}
    \caption{Qualitative results of AGM (left), PGM (middle) and PRM (right). The upper and lower line shows results without or with using the module, respectively. Best view in screen.}
    \label{fig:visual}
\end{figure}

\paragraph{Ablation study of AGM.}
To evaluate the efficacy of the Anchor Generation Module (AGM), we conducted a comparative analysis between randomly selected anchors and those generated using AGM. The results are presented in Tab.~\ref{three_schedule}. A comparison between the results in the first and second rows underscores the pivotal role played by AGM in polygon generation. The absence of AGM results in a performance decline of $56.0\%$ on IC15, $59.9\%$ on TotalText, and $47.3\%$ on CTW1500, respectively. AGM demonstrates its capability by generating a coarse anchor, providing the width and height of the text for the subsequent modules based on recognition confidence. The absence of these coarse anchors supplied by AGM renders the PGM incapable of generating precise boundary points. The qualitative results are illustrated in Fig.~\ref{fig:visual}.

\begin{table*}[t]
\centering
\renewcommand\arraystretch{1.2}
\setlength{\tabcolsep}{2mm}{
\begin{tabular}{cccccccccc}
\hline
\multirow{2}{*}{Attn. Threshold} & \multicolumn{3}{c}{ICDAR2015}   & \multicolumn{3}{c}{TotalText} & \multicolumn{3}{c}{CTW1500} \\ \cline{2-10} 
                                & P & R & H & P   & R  & H  & P   & R  & H \\ \hline
0.05 & 77.1 & 70.7 & 73.8  & 71.7 & 70.7 & 71.2 & 53.5 & 56.0 & 54.7 \\
0.04 & 77.1 & 70.6 & 73.7  & 71.7 & 70.7 & 71.2 & 53.6 & 56.1 & 54.8 \\
0.03 & 76.9 & 70.5 & 73.6  & 71.9 & 70.9 & 71.4 & 53.7 & 56.1 & 54.9 \\
0.02 & 76.9 & 70.5 & 73.6  & 71.9 & 70.9 & 71.4 & 53.8 & 56.3 & 55.0 \\
0.01 & 76.0 & 69.7 & 72.7  & 72.0 & 71.0 & 71.5 & 54.2 & 56.7 & 55.4 \\ \hline
\end{tabular}
}
\setlength{\abovecaptionskip}{0.13cm}
\caption{The influence of attention threshold. Attn. represents attention.}
\label{attn}
\end{table*}

\paragraph{Ablation study of PGM.}
To verify the effectiveness of the Polygon Generation Module (PGM), we conducted an ablation study where we removed the PGM and directly forwarded the coarse anchors to the Polygon Rectification Module. As illustrated in Tab.~\ref{three_schedule}, without PGM, the performance decreased by approximately $5.9\%$ on the IC15 dataset, $11.1\%$ on the TotalText dataset, and $5.2\%$ on the CTW1500 dataset, respectively. The qualitative results are illustrated in Fig.~\ref{fig:visual}. The PGM serves to transform the coarse anchors into boundary points that are in closer proximity to the text region, which plays a crucial role of the PGM in improving the quality of the boundary points. Notably, the PGM is supervised by the recognition loss without the detection annotations. 

\begin{table*}[h]
\centering
\renewcommand\arraystretch{1.2}
\setlength{\tabcolsep}{3.5mm}{
\begin{tabular}{cccccccccc}
\toprule
\multirow{2}{*}{Threshold} & \multicolumn{9}{c}{Results Generated from GT Points}  \\ \cline{2-10} 
                           & \multicolumn{3}{c}{ICDAR2015}  & \multicolumn{3}{c}{TotalText} & \multicolumn{3}{c}{CTW1500}   \\ \hline                  
IOU                       & P  & R  & H & P  & R  & H  & P  & R  & H  \\ \midrule
0.1 & 97.1 & 89.8 & 93.3 & 97.6 & 92.8 & 95.1 & 94.8 & 90.6 & 92.7 \\
0.3 & 95.7 & 88.6 & 92.0 & 91.4 & 86.6 & 88.8 & 80.0 & 76.4 & 78.1 \\
0.5 & 85.3 & 79.0 & 82.0 & 76.0 & 72.3 & 74.1 & 62.8 & 60.0 & 61.3 \\
0.7 & 48.4 & 44.8 & 46.6 & 32.9 & 31.3 & 32.1 & 27.9 & 26.7 & 27.3  \\ \hline
                           & \multicolumn{9}{c}{Results Generated from SPTS v2 Points}                                                   \\ \hline
                           
IOU                        & P  & R  & H  & P  & R  & H  & P  & R  & H  \\ \midrule
0.1           & 92.5     & 84.8  & 88.5   & 92.0      & 90.7   & 91.4  & 85.8     & 89.7  & 87.7   \\
0.3           & 90.0     & 82.5  & 86.1   & 87.9      & 86.7   & 87.3   & 71.8    & 75.0  & 73.4 \\
0.5           & 77.1     & 70.7  & 73.8   & 72.0      & 71.0   & 71.5  & 54.2     & 56.7  & 55.4  \\
0.7           & 32.8     & 30.1  & 31.4   & 27.3      & 26.9   & 27.1  & 23.3     & 24.4  & 23.8   \\ \hline
                           & \multicolumn{9}{c}{Results of Original SPTS v2}                                                        \\ \hline
                           
DIST                        & P  & R  & H  & P  & R  & H & P & R  & H \\ \midrule

5                          & 59.0     & 54.4  & 57.0   & 46.1      & 45.4   & 45.7   & 61.8     & 63.7  & 62.7\\
10                         & 88.4     & 80.3  & 84.2   & 70.0      & 69.1   & 69.5  & 74.3     & 76.5  & 75.7   \\
20                         & 92.7     & 84.2  & 88.2   & 82.7      & 81.6   & 82.1   & 83.0    & 85.4  & 84.2   \\
30                         & 94.0     & 85.4  & 89.5   & 87.8      & 86.6   & 87.2  & 85.3     & 87.9  & 86.6 \\ \bottomrule
\end{tabular}
}
\caption{The comparison involves SPTS v2 point detection and Point2Polygon predictions. 'GT Points' indicates polygons created from Ground Truth points, and 'SPTS v2 Points' denotes polygons derived from SPTS v2 predicted points.}
\label{result}
\end{table*}

\paragraph{Ablation study of PRM.}
We conducted a comparative analysis to evaluate the impact of the Polygon Rectification Module (PRM) in Tab.~\ref{three_schedule}. The PRM notably influences IC15, revealing that boundary points from PGM often extend beyond the boundaries in multi-oriented text. The qualitative results are illustrated in Fig.~\ref{fig:visual}. Without the PRM, the boundary points generated by the PGM tend to excessively broaden horizontally, extending well beyond the actual text region. However, the implementation of the PRM effectively eliminates these surplus points beyond the text's boundaries.

\paragraph{Ablation study of the attention threshold in PRM.}
Our method presets the attention threshold to remove the excess points beyond the text’s range. To evaluate the influence of different thresholds, we conduct experiments while maintaining the model structure unchanged, only modifying the threshold. The results are presented in Tab.~\ref{attn}. Notably, as the threshold increased, we observed a gradual improvement in the performance of IC15. Conversely, for Totaltext and CTW1500, we observed a slight decline in performance with higher attention thresholds. This suggests that there is a greater need for rectification in the case of multi-oriented text, which should follow high-quality standards.

\subsection{Quality of the Generated Polygon}
In this section, we conduct experiments to verify the quality of polygons generated by our methods. There are two types of points considered: 1) Center points generated based on the Ground Truth (GT Points) of the testing set; and 2) Center points predicted using SPTS v2 (SPTS v2 Points). Subsequently, we generate polygons based on these two types and evaluate their performance. It is important to note that the accuracy of polygons generated from SPTS v2 is constrained by its performance. 

We test the performance across four IOU thresholds. As illustrated in Tab.~\ref{result}, under 0.5 IOU threshold, polygons generated from GT Points achieve $82\%$ on IC15, $74.1\%$ on TotalText, and $61.3\%$ on CTW1500, respectively, in terms of Hmean; 
while the polygons generated from SPTS v2 Points achieve $73.8\%$ on IC15, $71.5\%$ on TotalText, and $55.4\%$ on CTW1500, respectively. We can also observe that as the IOU threshold decreases, the performance gap between SPTS v2 Points and GT Points narrows.

To further evaluate the performance of the generated points from SPTS v2, we also directly compare the detection performance with SPTS v2. The term ``DIST'' denotes the number of pixels within which predicted points are considered accurate relative to ground truth points. To relate IOU to DIST, we initially calculate the difference in performance between polygons generated from SPTS v2 Points and GT Points on IC15 under a 0.5 IOU. Using SPTS v2 Points achieves $89.5\%$ of the performance achieved using GT Points. Therefore, at an IOU of 0.5, the corresponding DIST is about 30. Mapping this to IC15, the generated polygons from SPTS v2 Points achieve $82.5\%$ of SPTS v2's performance. 
The results demonstrate our method can be effectively integrated into an existing single-point text spotter to produce polygon results without increasing the labeling costs.

\begin{figure*}[!t]
    \centering
    \includegraphics[width=0.9\linewidth]{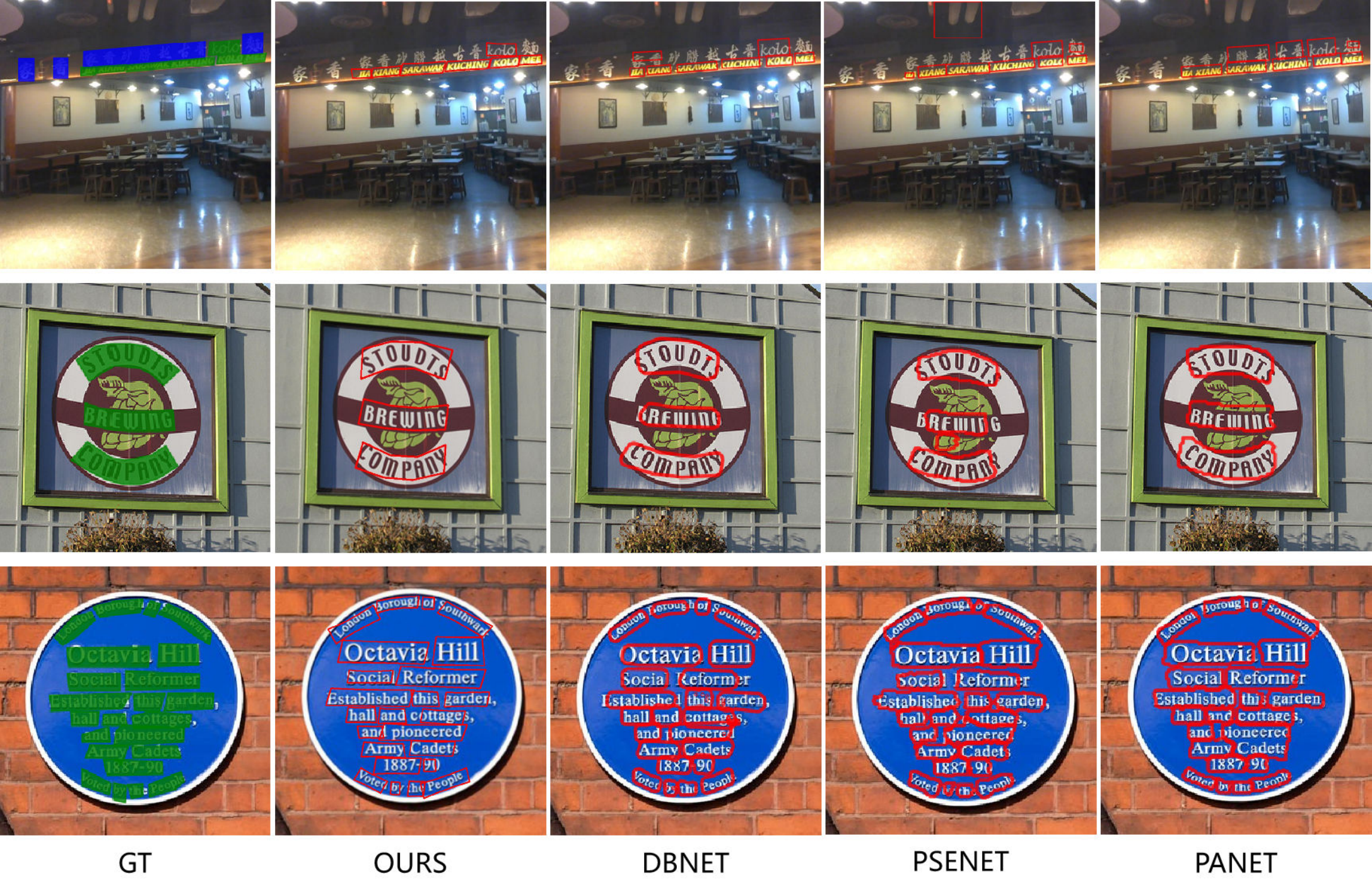}
    \caption{ Visualization results of training using the generated polygons. The green-filled polygons are the ground truths, the blue-filled polygons are the ground truths that are marked as ``don't care'', and the red polygons are the detection results. The second column represents the polygon generated by our method in combination with the existing single point text detector (SPTS v2). The third to sixth columns represent the effect of different detectors trained using the generated polygons. Best view in screen.
    }
    \label{fig:visual_det}
\end{figure*}

\begin{table*}[!t]
\centering
\renewcommand\arraystretch{1.2}
\setlength{\tabcolsep}{3mm}{
\begin{tabular}{cccccccccc}
\hline
\multirow{2}{*}{Model} & \multicolumn{3}{c}{ICDAR2015}  & \multicolumn{3}{c}{TotalText} & \multicolumn{3}{c}{CTW1500} \\ \cline{2-10} 
                       & P  & R  & H & P  & R & H & P & R  & H  \\ \hline
& \multicolumn{9}{c}{Using Original Ground Truth} \\ \hline
DBNet + resnet50        & 89.8     & 79.4  & 84.3 & 84.5  & 79.6   & 82.0  & 68.3  & 63.4  & 65.8 \\
PSENET + resnet50       & 83.6     & 76.2  & 79.7 & 82.3  & 77.2   & 79.6  & 77.5  & 82.9  & 80.1 \\
PANET + resnet18        & 83.1     & 72.9  & 77.6 & 85.9  & 77.3   & 81.4  & 86.3  & 82.0  & 84.1 \\ \hline
& \multicolumn{9}{c}{Using Polygons Generated from SPTS v2 Point}                                                   \\ \hline
DBNet + resnet50        & 82.4     & 67.9  & 77.5 & 81.8  & 65.0   & 72.4  & 61.5  & 42.8  & 50.5 \\
PSENET + resnet50       & 77.4     & 54.2  & 63.8 & 76.5  & 55.8   & 64.5  & 56.3  & 41.7  & 47.9 \\
PANET + resnet18        & 67.8     & 51.1  & 58.3 & 77.9  & 55.9   & 65.1  & 67.2  & 42.9  & 52.3 \\ \hline
& \multicolumn{9}{c}{Using Polygons Generated from GT Point}
                \\ \hline
DBNET + resnet50            & 87.4      & 65.5   & 74.9  & 84.6        & 65.9    & 74.1   & 62.9       & 44.1   & 51.8  \\
PSENET + resnet50           & 72.7      & 58.2   & 64.6  & 76.7        & 56.9    & 65.3   & 56.7       & 38.6   & 46.0  \\
PANET + resnet18            & 77.2      & 51.8   & 62.0  & 79.1        & 57.4    & 66.5   & 58.0       & 45.5   & 51.0  \\ \hline
\end{tabular}
}
\caption{Detection results using the polygon generated from our method. All experiments are conducted in the mmocr. The Raw Ground Truth means using the polygon from Ground Truth to train the detector. SPTS v2 Point Generated Annotation involves utilizing points from SPTS v2 to generate polygons for training the detector. GT Point Generated Annotation entails using points from the ground truth (GT) to generate polygons for training the detector.}
\label{prediction_gt}
\end{table*}

\subsection{Evaluating on Existing Detector}
To facilitate a comprehensive comparison between the polygons generated by our method and the ground truth polygons, we utilized these polygons for training the DBNet~\cite{liao2020real}, PANet~\cite{wang2019efficient}, and PSENet~\cite{wang2019shape} models. Specifically, we employed quadrilaterals as ground truth annotations for the IC15 dataset, polygons for the TotalText dataset and the CTW1500 dataset. All training procedures were conducted within the mmocr framework~\cite{kuang2021mmocr}, and the results are outlined in Tab.~\ref{prediction_gt} with an IoU threshold of 0.5. The results are categorized into two types: 1) GT Points Generated Annotation, where we generated the center points based on the ground truth (GT Points) from the training set, and used these polygons to train the detection model; 2) SPTS v2 Points Generated Annotation, where we utilized the SPTS v2 to predict the center points (SPTS v2 Points), and employed the resulting polygons for training the detection model. 

By employing the SPTS v2 Points Generated Annotation for training the detector, the DBNet demonstrates notable performance with an accuracy of $77.5\%$ on IC15, $72.4\%$ on TotalText, and $50.5\%$ on CTW1500. Similarly, the PSENet achieves $63.8\%$ on IC15, $64.5\%$ on TotalText, and $47.9\%$ on CTW1500, while the PANet attains $58.3\%$ on IC15, $65.1\%$ on TotalText, and $52.3\%$ on CTW1500.
Furthermore, it is evident from the results that SPTS v2 Points Generated Annotation achieves approximately $86\%$ of the performance observed when using polygons from the ground truth to train the model. This highlights the effectiveness of our method. 
Qualitative results are shown in the Fig.~\ref{fig:visual_det}. These images are not included in the training set.

Moreover, by comparing the seventh row of Tab.~\ref{result} with the fourth row of Tab.~\ref{prediction_gt}, we observe that the results of DBNet using polygons generated from SPTS v2 Point outperform the performance achieved with polygons without training by detectors, with  $3.7\%$ and $0.9\%$ improvement on IC15 and Totaltext, respectively, in terms of Hmean. It demonstrates that while the generated polygon may not achieve perfect accuracy, its overall quality remains high. This high-quality polygon allows the detector to further exceed the performance of the polygon generated from SPTS v2 Point.
\section{Discussion}

\subsection{Visualization}
\begin{figure}[!t]
    \centering
    \includegraphics[width=8.5cm, height=9cm]{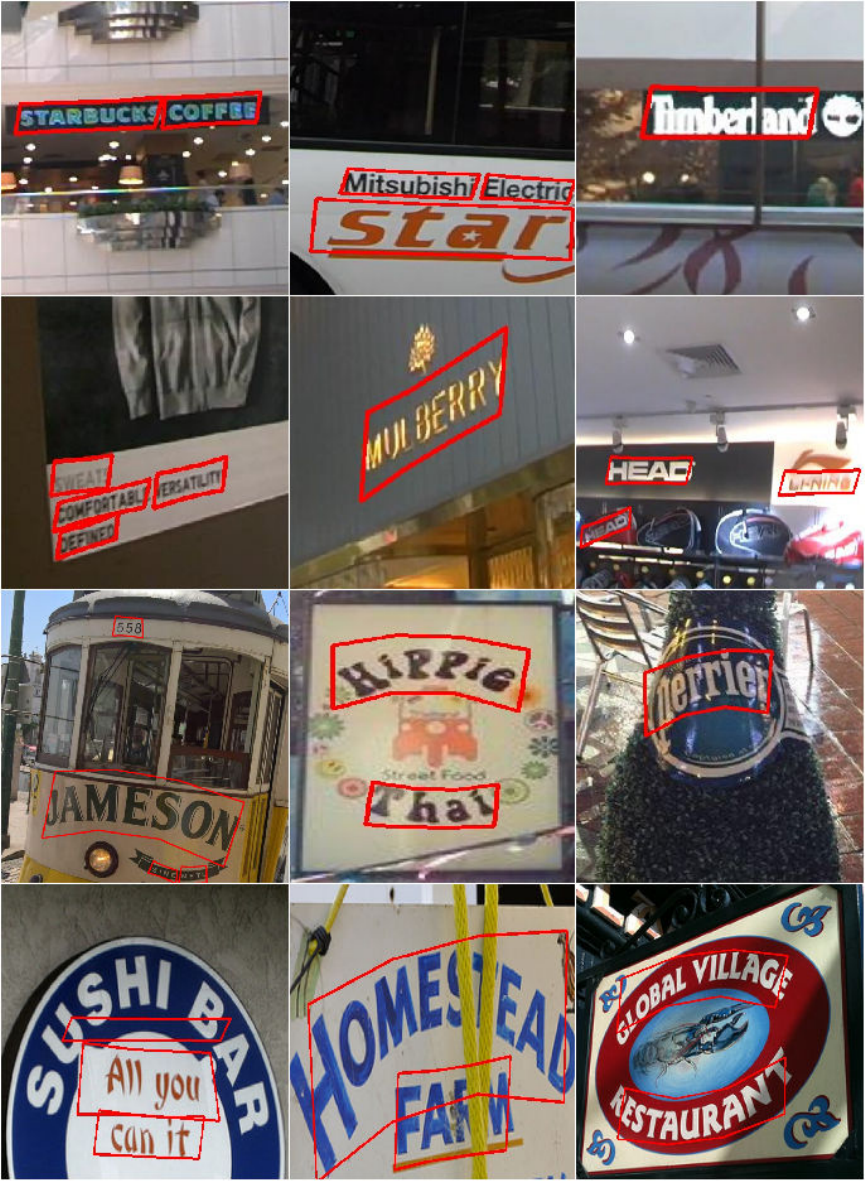}
    \caption{Visualization results of Point2Polygon. The first row shows horizontal examples. The second row shows tilted examples. The third row shows slightly curved examples. The fourth row shows some failure examples. Best view in screen.
    }
    \label{fig:discuss_vis}
\end{figure}
Example visualization results of the polygon generated by Point2Polygon are presented in Fig.~\ref{fig:discuss_vis}. 
Point2Polygon can generate visually-plausible polygons for various text including horizontal, multi-oriented, and arbitrarily-shaped scene text. We also observe it shows robust performance in generating polygons for slightly curved text.

\subsection{Limitation}
A noted limitation is that the generated polygon does not entirely align with the Ground Truth, especially for some highly curved texts, as shown in the last line of Fig.~\ref{fig:discuss_vis}. Nevertheless, we believe that the current level of precision has the potential to alleviate the annotation labor and serve as a foundation to drive future studies in this domain. 
Additionally, studies such as TTS \cite{kittenplon2022towards}, SPTS \cite{liu2023spts}, and TOSS \cite{tang2022you} have shown that without any specific location information but with transcription-only, successfully training a robust scene text spotter is still viable. This suggests that spotters are able to implicitly learn text locations in images. However, Point2Polygon currently requires a reference point to generate the polygon, indicating room for potential enhancement.

\begin{table*}[!t]
\centering
\renewcommand\arraystretch{1.4}
\setlength{\tabcolsep}{4mm}{
\begin{tabular}{cccccccccc}
\hline
\multirow{2}{*}{Model} & \multicolumn{3}{c}{ICDAR2015}  & \multicolumn{3}{c}{TotalText} & \multicolumn{3}{c}{CTW1500} \\ \cline{2-10} 
         & P        & R        & H     & P        & R        & H       & P       & R       & H       \\ \hline
TESTR    & 90.3     & 89.7     & 90.0    & 93.4     & 81.4     & 86.9    & 92.0      & 82.6    & 87.1    \\
DeepSolo & 92.8     & 87.4     & 90.0    & 93.9     & 82.1     & 87.6    & -       & -       & -       \\
ours     & 85.3     & 79.0       & 82.0    & 76.0       & 72.3     & 74.1    & 62.8    & 60.0      & 61.3    \\ \hline
\end{tabular}
}
\setlength{\abovecaptionskip}{0.2cm}
\caption{Compared with DETR-based methods, they are supervised by polygons.}
\label{detr}
\end{table*}

We set multiple anchors to crop the original image, which is inefficient. In the future, we will try to perform region selection on the feature map to improve efficiency, and explore parallelization methods to improve the running speed of our model

\subsection{Compared with DETR-based methods}
The results of our method compared with the DETR-based methods are shown in the Tab.~\ref{detr}.  We mainly compare with DeepSolo \cite{ye2023deepsolo} and TESTR \cite{zhang2022text} . DeepSolo and TESTR are supervised by polygon annotations and have been trained on a large amount of real data. In contrast, our model is only annotated with single points and trained for recognition only on synthetic data. Therefore, it is reasonable that there is a certain gap between our method and the other methods in the table. In the future, we will continue to explore more optimal methods to reduce the gap with fully supervised methods.

\section{Conclusion}
We introduce the Point2Polygon for
effectively evolve points into polygons from coarse to fine through multi-granularity recognition information. A key aspect of our method is the use of synthetic datasets for recognition information, effectively bypassing the need for any manual annotation beyond a single point. Our extensive experiments demonstrate the strength and reliability of our method. These tests involve direct comparisons to ground truth data, applying our approach in conjunction with existing text detection systems, and assessing its performance when integrated with the single-point spotting method. 
While there is a noticeable distinction between our generated results and actual ground truths, Point2Polygon has shown encouraging results. Additionally, it is worth noting that this method may aid in automatically producing more detailed information in large-scale image-text pairs, thereby assisting in extensive training.
We believe this approach not only demonstrates significant potential in the realm of point-to-polygon evolution but also sets a strong baseline for future research in this area.

{
    \small
    \bibliographystyle{ieeenat_fullname}
    \bibliography{main}
}

\end{document}

%% file: preamble.tex
\usepackage[dvipsnames]{xcolor}
